\documentclass{article}

\usepackage{PRIMEarxiv}

\usepackage[utf8]{inputenc} % allow utf-8 input
\usepackage[T1]{fontenc}    % use 8-bit T1 fonts
\usepackage{hyperref}       % hyperlinks
\usepackage{url}            % simple URL typesetting
\usepackage{booktabs}       % professional-quality tables
\usepackage{amsfonts}       % blackboard math symbols
\usepackage{nicefrac}       % compact symbols for 1/2, etc.
\usepackage{microtype}      % microtypography
\usepackage{lipsum}
\usepackage{fancyhdr}       % header
\usepackage{graphicx}       % graphics
\usepackage{amsmath}
\usepackage{amssymb}
\usepackage{algorithm,algpseudocode}

\usepackage{wrapfig}
\usepackage{multirow}
\usepackage{capt-of}
\usepackage{caption2}
\graphicspath{{media/}}     % organize your images and other figures under media/ folder

\title{Semi-supervised Anomaly Detection Algorithm based on KL divergence (SAD-KL)
}

\author{
  Chong Hyun Lee \\
  Department of Ocean System Engineering,\\
  Jeju National University \\
  \texttt{chonglee@jejunu.ac.kr} \\
   \And
  Kibae Lee \\
  Department of Ocean System Engineering, \\
  Jeju National University \\
  \texttt{kibae0211@stu.jejunu.ac.kr} \\
}

\begin{document}
\maketitle

\begin{abstract}
The unlabeled data are generally assumed to be normal data in detecting abnormal data via semi-supervised learning. This assumption, however, causes inevitable detection error when distribution of unlabeled data is different from distribution of labeled normal dataset. To deal the problem caused by distribution gap between labeled and unlabeled data, we propose a semi-supervised anomaly detection algorithm using KL divergence (SAD-KL). The proposed SAD-KL is composed of two steps: (1) estimating KL divergence of probability density functions (PDFs) of the local outlier factors (LOFs) of the labeled normal data and the unlabeled data (2) estimating detection probability and threshold for detecting normal data in unlabeled data by using the KL divergence. We show that the PDFs of the LOFs follow Burr distribution and use them for detection. Once the threshold is computed, the SAD-KL runs iteratively until the labeling change rate is lower than the predefined threshold. Experiments results show that the SAD-KL shows superior detection probability over the existing algorithms even though it takes less learning time.
\end{abstract}

\section{Introduction}
The purpose of anomaly detection is to detect abnormal samples that deviate from the predefined normality [1]. Anomaly detection has various applications in medicine, security, and manufacturing [1]. The unsupervised anomaly detection algorithms assume that most of the samples are normal and learn features of normal samples [2, 3, 4, 5, 6]. Anomaly detection algorithms named as the one class support vector machine (OC-SVM) [2] and support vector data description (SVDD) [3] are reported. The deep learning approaches have shown outstanding performance by successively learning high-dimensional data [4, 5, 6]. The Deep SVDD [4], Deep Multi-sphere SVDD [5] and deep robust one-class classification (DROCC) [6] have been proposed to learn representation of normal samples. In spite of improved performance, these unsupervised approaches have limited detection performance since they do not train abnormal information in dataset.

Some labeled data, as well as unlabeled data, may be utilized and especially small number of anomalous samples in labeled data can be used for training. Song et al. [7] and Akvay et al. [8] proposed semi-supervised anomaly detection models which use reliable normal samples in training of unlabeled data. Nevertheless, they have limited performance because these models do not train the abnormalities yet. Ruff et al. [9] proposed a deep semi-supervised anomaly detection (Deep SAD) that learns anomalous samples in labeled data by assuming that most unlabeled samples are normal. The Deep SAD moves the normal data to the center of the latent space and labeled abnormal samples away from the center. An unsupervised outlier exposure (OE) approach for learning OE data and unlabeled data has been proposed by using similar assumption of semi-supervised anomaly detection [10]. The algorithm based on unsupervised OE learning trains a binary classifier by using OE data, the unlabeled abnormal and normal data. The hyper sphere classification (HSC) [11] algorithm based on unsupervised OE learning, uses the relative distance in latent space for training. Even the Deep SAD and HSC are advanced algorithms using abnormal samples or OE data in the training process, they cannot overcome performance degradation as the number of abnormal samples in the unlabeled dataset increases. To overcome the degradation, semi-supervised algorithm that can use both normal and abnormal data in unlabeled dataset, algorithm using probabilistic labeling (SAD-PL) is presented [12]. The SAD-PL shows superior detection performance over previous algorithms including Deep SAD and HSC by using probabilistic labeling based on Neyman-Pearson (NP) detection criterion. Despite of outstanding detection performance, the SAD-PL has weakness of long training time to obtain the best detection probability and performance dependency on statistics of unlabeled data.

To reduce the detection dependency on distribution gap between labeled and unlabeled data, we propose a detection algorithm using KL divergence. The KL divergence is computed by using two local outlier factors (LOF) scores obtained from both labeled and unlabeled data, respectively. We show that PDFs of LOF scores follow the Burr distribution. The SAD-KL uses estimated parameters of the Burr distributions for detection.

\section{Related work}
\paragraph{One Class Classification}
Typically, anomaly detection task is treated as an unsupervised learning problem assuming that most training data are normal. The one class classification aims to lean a model that accurately describes “normality”. The most representative kernel-based method for one class classification is the OC-SVM [2]. The objective of the OC-SVM finds a maximum margin hyperplane in feature space. On the other hand, the SVDD [3] attempts to separate data into hypersphere instead of hyperplane. The objective of SVDD is to find smallest hypersphere that encloses the data in feature space. Classical anomaly detection methods including OC-SVM and SVDD are limited in their scalability to large datasets. Recent deep learning approaches have shown outstanding performance by overcoming the problems which cannot solve by classical algorithms. The Deep SVDD [4], a representative deep approach, trains a neural network by minimizing the volume of a hypersphere that encloses the network representations of the data. This minimizing of the volume of the hypersphere forces the network to map the data to the center in latent space. For learning “normality” of complex data, deep one class classification algorithms such as Deep SVDD [4], Deep Multi-sphere SVDD [5] and DROCC [6], have been reported.

\paragraph{Semi-supervised Anomaly Detection based on deterministic labeling}
Unlike the assumption of the typical anomaly detection task, some labeled data includes a small fraction of anomalous samples that can be used in real-world applications. The hybrid SAD [7] and GANomaly [8] learn reliable normal samples in unlabeled data as well as labeled data. However, these methods have limitations in detecting anomalies since they do not learn the anomalous samples. The Deep SAD [9] related to Deep SVDD is proposed for semi-supervised anomaly detection. By assuming that the majority of the unlabeled data consists of normal samples, the Deep SAD learns by forcing the normal data to be concentrated and anomalous samples in labeled data to move away from the center of the latent space. The HSC [11] is a classification algorithm based on unsupervised OE learning that uses the relative distance in latent space for training the model. The HSC can be modified as semi-supervised anomaly detection method by replacing OE data with labeled data. Semi-supervised anomaly detection approaches including Deep SAD and HSC find a compact description of the normality while also discriminating the labeled anomalies. However, the assumption of most unlabeled samples being normal inevitably causes performance degradation as the number of abnormal samples in the unlabeled dataset increases.

\paragraph{Semi-supervised Anomaly Detection based on probabilistic labeling}
The positive and unlabeled metric learning for anomaly detection (PUMAD) [13] learns reliable normal samples and potential anomalies among the unlabeled data using triplet loss. This algorithm identifies normal and abnormal samples in unlabeled data with distance hashing-based method by assuming that anomalies are clustered in latent space. The log-based anomaly detection via probabilistic label estimation (PLELog) [14] learns unlabeled data with the stability score of each cluster via hierarchical density-based spatial clustering of application with noise (HDBSCAN) [15]. These methods, however are only applicable to group or collective anomalies. The SAD-PL [12] can handle both group and collective anomalies by using LOF score and learning unlabeled data with estimated probabilistic labels obtained by the NP criterion. However, the SAD-PL employs ensemble networks to find the best threshold satisfying the give detection probability so that it requires heavy computation. In addition, because the SAD-PL estimates the threshold by using labeled normal data, its detection performance may degrade when distribution of unlabeled data is different from distribution of labeled normal dataset.

\section{SAD-KL}
In this section, we introduce the SAD-KL along with the existing semi-supervised algorithms of SAD-PL as shown in Figure 1. The proposed SAD-KL uses feature representations $\phi(\mathbf{x})$ through autoencoder pretraining and learns the encoded normal samples to close centroid $\mathbf{c}$ in latent space without the decoding network.Then, the SAD-KL is trained according to the estimated KL divergence, which uses the PDF of $s_{r}$ from labeled normal and the PDF of $s _{u}$ from unlabeled data among $s=\rm{LOF}_{k}(\phi(x))$. The LOF scoring based on the relative distance between k neighbor samples is known to show robust performance in the multimodal normality case [16]. As shown in Figure 1, the s of the object p is computed as follows: (1) computing k-distances of p with the k neighbor samples $o_{i}$ (2) computing k-distances of $o_{i}$ with the k neighbor samples (3) computing ratio of average of the k-distances obtained by (2) to the k-distances by (1). To detect a small number of group anomaly samples, the SAD-KL sets k large enough to cover the relative distance between normal and abnormal samples.

\begin{figure}
  \centering
  \includegraphics[width=\textwidth]{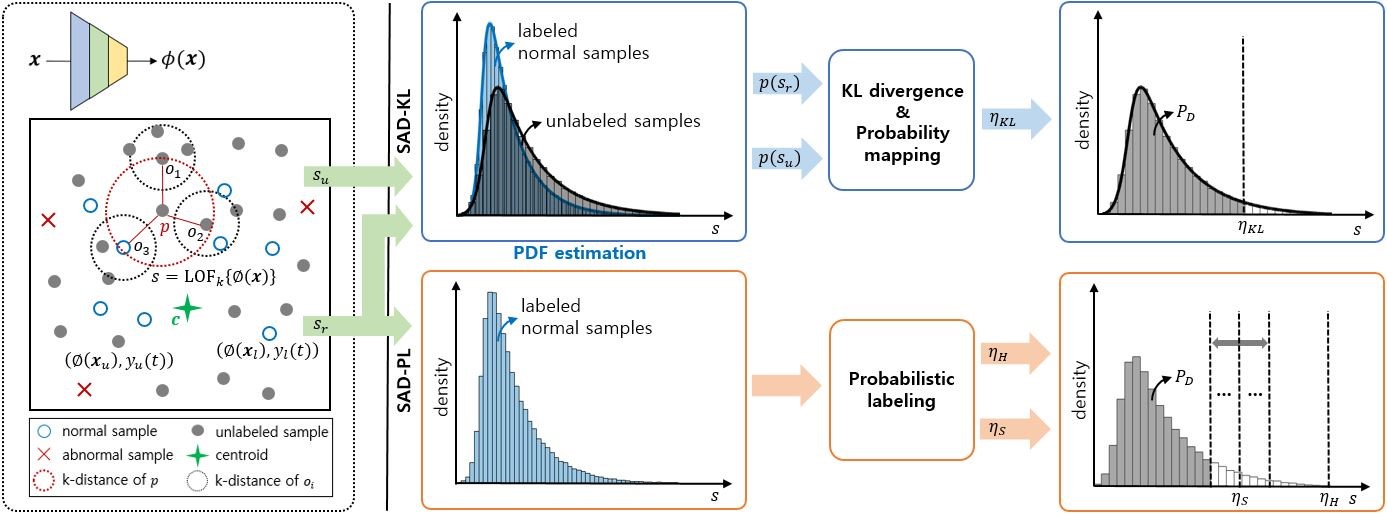}
  \vspace*{-5mm}
  \caption{Overview of the proposed SAD-KL and existing SAD-PL algorithms.}
  \label{fig:fig1}
  \vspace*{-5mm}
\end{figure}

The NP criterion is used to determine the threshold that satisfies the detection probability under a given constraint [17]. The threshold $\eta$ is computed by using the detection probability $P_{D}$ as

\begin{equation}
P_{D}=\int_0^\eta p(s|H_{0})ds
\end{equation}

where $p(s \vert H_{0})$ denotes the PDF of $s$ obtained from data under condition $H_{0}$. Note that the $P_{D}$ depends on changes of threshold $\eta$. Therefore, when the distribution of unlabeled data is almost the same as the distribution of labeled normal data, threshold $\eta$ obtained from labeled normal data can gives almost the same detection probability on the unlabeled data. However, if unlabeled data contains diverse normal samples so that it has different distribution, then same detection probability $P_{D}$ cannot be guaranteed with the threshold $\eta$ pre-obtained from the labeled data. To consider the distribution difference in computing $P_{D}$, we use KL divergence and adopt exponential function mapping the obtained KL divergence to probability value. Consequently, a detection probability formula for the unlabeled data is proposed as follows:

\begin{equation}
P_{D}=exp({\frac{KL(P||Q)}{\beta}})
\end{equation}

where $KL(P \vert \vert Q)=\int_{-\infty}^\infty (s)\log{p(s)/q(s)}$denotes the KL divergence of PDF $p(s)$ of labeled normal data and $q(s)$ of unlabeled data and the $\beta$ is a parameter to be determined. Since the $KL(P \vert \vert Q)$ represents difference between the labeled and the unlabeled data distributions, the detection probability of normal data in unlabeled data can be estimated by proper choice of $\beta$. Once appropriate $\beta$ is chosen, the threshold $\eta$ can be found by setting (1) equals to (2).

As shown in the following simulations, the $\beta$ is not insensitive to dataset so that sole choice of $\beta=2500$ is sufficient to have the best detection performance on three datasets. Furthermore, the PDF of $s$ can be assumed as Burr distribution so that we can use the following PDF $p(s)$ and CDF $F(s)$ [18, 19]:

\begin{equation}
p(s)={\frac{kcs^{c-1}}{((1+s)^{c})^{k+1}}}
\end{equation}

\begin{equation}
F(s)=1-(1+s^{c})^{-k}
\end{equation}

where $c$ and $k$ are shape parameters that can cover a broad set of skewness and kurtosis [18, 19]. The threshold $\eta$ can be computed by finding s satisfying $P_{D}$ in (2) as follows:

\begin{equation}
\eta=((1-P_{D})^{-{\frac{1}{k}}}-1)^{\frac{1}{c}}
\end{equation}

\algnewcommand\INPUT{\item[\textbf{Input:}]}%
\algnewcommand\OUTPUT{\item[\textbf{Output:}]}%

\begin{wrapfigure}{R}{0.45\textwidth}
\raisebox{0pt}[\dimexpr\height-1\baselineskip\relax]{
\begin{minipage}{0.45\textwidth}
\begin{algorithm}[H]
\caption{Learning procedure of SAD-KL}
\begin{algorithmic}[1]
\INPUT Labeled data $(\mathbf{x}_{l}, y_{l})$, Unlabeled data $\mathbf{x}_{u}$, Number of neighbors $k$, Threshold for labeling change rate $\epsilon$
\OUTPUT Trained model $\phi$
\State \textbf{Initialize:}
\State\hspace{\algorithmicindent} Pretrain autoencoder: $\phi$
\State\hspace{\algorithmicindent} Compute centroid: $\mathbf{c}$
\State \textbf{Determine detection probability:}
\State\hspace{\algorithmicindent} Compute LOF score: $s=\mathrm{LOF}_{k}(\phi(\boldsymbol{x}))$
\State\hspace{\algorithmicindent} Estimate PDFs: $p(s)$, $q(s)$
\State\hspace{\algorithmicindent} Compute detection probability: $P_{D}$
    \For{$t=1, 2, \dotsb$}
        \State Compute LOF score: $s=\mathrm{LOF}_{k}(\phi(\boldsymbol{x}))$
        \State Estimate PDF: $q(s) \rightarrow c, k$
        \State Compute threshold: $\eta$
        \State Probabilistic label: $y_{u}(t)$
        \State Train model: $\phi$
        \State \textbf{if} $\Delta y_{u}(t)<\varepsilon$ \textbf{then break}
    \EndFor
\end{algorithmic}
\end{algorithm}
\end{minipage}
}
\end{wrapfigure}

Note that $\eta$ changes because the PDF of $s$ varies as SAD-KL learns. Once $\eta$ is estimated by using equation (2), the label $y(t)$ at $t$ can be found as follows in [12]. The labels $y_{l}(t)$, $l \ge n$ for $n$ labeled samples $x_{l}$ and the labels $y_{u}(t)$, $u > n$ for unlabeled data $x_{u}$, are as follows:

\begin{equation}
y_{l}(t) = \begin{cases}
1, \quad \forall t \quad \text{for normal} \\
0, \quad \forall t \quad \text{for abnormal}
\end{cases}
\end{equation}

\begin{equation}
y_{u}(t) = \begin{cases}
P_{D}, \quad s_{u} \le \eta \\
1-P_{D}, \quad s_{u} > \eta
\end{cases}
\end{equation}

\setcounter{figure}{1}
\begin{figure}
  \centering
  \includegraphics[width=\textwidth]{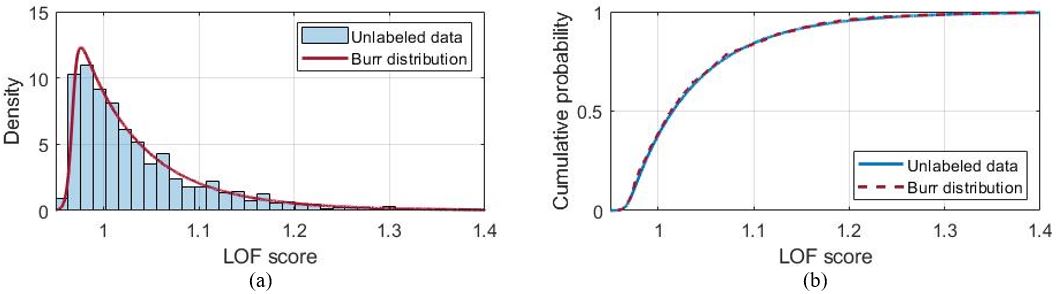}
  \vspace*{-5mm}
  \caption{(a) Histogram, the Burr PDF and (b) the Burr CDF of the $s_{u}$.}
  \label{fig:fig2}
\end{figure}

\setcounter{figure}{2}
\begin{figure}
  \centering
  \includegraphics[width=\textwidth]{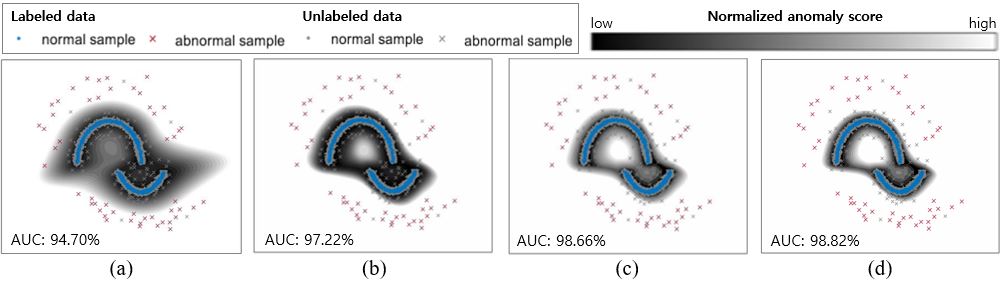}
  \vspace*{-5mm}
  \caption{The decision boundaries and test AUCs of (a) Deep SAD, (b) SAD-PL (Hard), (c) SAD-PL (Soft), and (d) SAD-KL.}
  \label{fig:fig3}
\end{figure}

where $s_{u}$ represents the LOF score of $x_{u}$. Note that $y_{u}(t)$ implies the probabilistic label for the unlabeled $x_{u}$. For network $\phi$ training, the SAD-KL uses datasets {$x_{l},y_{l}(t)$  ,$l \le n$} and {$x_{u}$,$y_{u}(t)$,$u>n$}, which are composed of n labeled and $(m-n)$ unlabeled data from the corresponding probabilistic labels estimated in (6) and (7). The SAD-KL use the same objective function of the SAD-PL described as follows:

\begin{equation}
\min_{W} \, \frac{1}{m} \sum\limits_{i=1}^{m} y_{i}(t)d(\phi(\mathbf{x}_{i})) + (1-y_{i}(t))(1-d(\phi(\mathbf{x}_{i})))+\frac{\lambda}{2} \sum\limits_{j=1}^{J}\Vert \mathbf{W}^{j} \Vert_{F}^{}
\end{equation}

where $\mathbf{W}^{j}$ is the weights of layer $j \in \{ 1, \dotsb, J \}$, $\Vert \cdot \Vert_{F}$ denotes the Frobenius norm and $d(\phi(\mathbf{x}_{i}))$ is the function of the distance from c using Geman-McClure loss defined as

\begin{equation}
d(\phi(\mathbf{x}_{i}))=\frac{\Vert \phi(\mathbf{x}_{i})-\mathbf{c} \Vert ^{2}}{\Vert \phi(\mathbf{x}_{i})-\mathbf{c} \Vert ^{2} + 1}
\end{equation}

The properties of the proposed SAD-KL follow those of SAD-PL [12] and runs until the labeling change rate $\triangle y_{u}(t)$ at iteration $t$ and is lower than the preset threshold $\epsilon$. The $\triangle y_{u}(t)$ is computed as

Algorithm 1 shows the overall learning procedure of the SAD-KL.

\begin{equation}
\Delta y_{u}(t)=\frac{1}{m-n} \sum_{u=n+1}^{m}\left|\frac{y_{u}(t)-y_{u}(t-1)}{2 P_{D}-1}\right|
\end{equation}

\setcounter{figure}{3}
\begin{wrapfigure}{R}{0.5\textwidth}
\centering
\raisebox{0pt}[\dimexpr\height+1\baselineskip\relax]{%
        \includegraphics[width=0.45\textwidth]{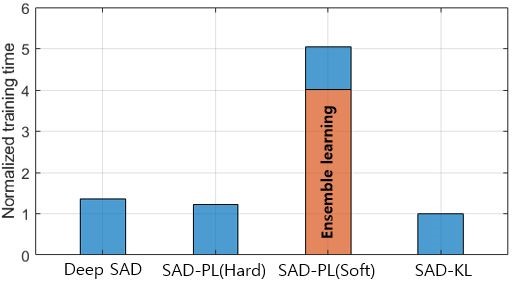}%
    }%
\caption{The training times normalized by of SAD-KL.}
  \label{fig:fig4}
\end{wrapfigure}

\setcounter{figure} {4}
\begin{figure}[b]
  \centering
  \includegraphics[width=\textwidth]{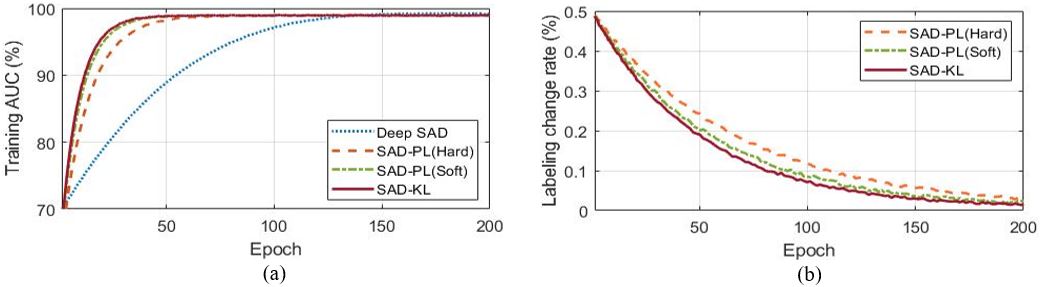}
  \vspace*{-5mm}
  \caption{(a) Training AUC and (b) labeling change rate of anomaly detection models according to epoch.}
  \label{fig:fig5}
  \vspace*{+12mm}
\end{figure}

We compare the anomaly detection models described above with two moons dataset. The training data consist of a total of 10000 samples, of which 10$\%$ are labeled data and 90$\%$ are unlabeled data. Five percent of labeled data and 1$\%$ of unlabeled data consist of abnormal samples. The test data consists of 1000 normal and abnormal samples. The normal samples contain Gaussian noise with a variance of 0.3 in big moon and small moon patterns. The abnormal samples are generated with a uniform distribution. The models for comparison use the same encoding network of the pretrained autoencoder. The encoding network consists of two hidden layers of 100 nodes, followed by ELU activations that represent a two-dimensional input in a two-dimensional latent space. For autoencoder pretraining, we employ the above architectures for the encoding networks and then construct decoding networks symmetrically. We set $\zeta=1$ for the Deep SAD, and $k=100$ and $\epsilon=0.0001$ for the SAD-PL and the SAD-KL. For training of SAD-PL with soft labeling, we use $Q=20$ by setting $P_{M}=0$ to $P_{M}=0.1$ with an $\delta=0.05$ interval. To determine $P_{D}$in SAD-KL learning, we set $\beta=500$. We set $\lambda=10^{-6}$ and use the Adam optimizer with a learning rate of $10^{-5}$ and train the Deep SAD with 200 epochs.

We plot the histogram, the estimated PDF and the CDF of $s_{u}$ of unlabeled data in Figure 2, which validates the Burr distribution assumption. Figure 3 shows the decision boundaries with training data and the test AUC (area under the receiver operating characteristic curve) of anomaly detection models. The decision boundaries in Figure 3 are expressed by an upper bound on 10$\%$ of the anomaly score normalized via min-max scaling. The anomaly score is computed $\|\phi(x)-c\|^{2}$ for all models. The training times normalized by SAD-KL is shown in Figure 4. As shown in Figure 3, the SAD-KL and SAD-PL represent the tighter decision boundaries than the Deep SAD which assumes that all unlabeled data are normal. Even the SAD-KL has tight decision boundaries and high AUC of 98.82$\%$, it takes the least training time as shown in Figure 4. Figure 5 shows the training AUCs and the label change rate $\Delta y_{u}(t)$ of the comparative models according to epoch. As shown in of Figure 5, the proposed SAD-KL shows high learning efficiency and faster convergence of $\Delta y_{u}(t)$ due to KL divergence based efficient learning whereas the Deep SAD shows slower learning.

\section{Experiments}
We evaluate the proposed SAD-KL on the well-known MNIST [20], CIFAR10 [21], and MNIST-C [22] datasets. The SAD-KL is compared with the methods based on one class classification. We present results from unsupervised methods of SVDD [3] and the Deep SVDD [4] and semi-supervised methods of the Deep SAD [9], HSC [11] and SAD-PL [12]. We implement the semi-supervised HSC by replacing OE data with the labeled data. We run all experiments for $\nu\in\{0.1, 0.25, 0.5\}$ of SVDD with a Gaussian kernel and show the corresponding results. The deep models use the same encoding network structure in the pretrained autoencoder. We employ LeNet-type convolutional neural networks (CNNs), where each convolutional module consists of a convolutional layer followed by leaky ReLU activations and 2×2 max-pooling. In the MNIST and MNIST-C experiments, we employ a CNN with two modules, 8×(5×5) filters followed by 4×(5×5) filters, and a final dense layer of 32 units. In the CIFAR10 experiments, we employ a CNN with three modules, 32×(5×5) filters, 64×(5×5) filters, and 128×(5×5) filters, followed by a final dense layer of 128 units. For the pretraining autoencoder, we employ identical encoding networks and then construct the decoding networks symmetrically, where we replace max-pooling with simple upsampling and convolutions with deconvolutions. We use a batch size of 200 and set $\lambda=10^{-6}$. We also use the Adam optimizer with a learning rate of $10^{-5}$. For experiments on the Deep SVDD, we use the one-class Deep SVDD model [4]. We run all experiments for $\zeta\in\{0.01,0.1,1,0,100\}$ of the Deep SAD and show the best results. We set $k=200$ and stop learning when $\triangle y_{u}(t)$ is less than $\epsilon=0.001$ for both SAD-KL and SAD-PL. The SAD-PL is evaluated via two separate hard and soft labels. In soft labeling, we set $Q=101$ by setting $P_{M}=0$ to $P_{M}=0.2$ with an $\delta=0.002$ interval and select the $P_{M}(o)$ in which $\triangle s$ is maximum in $T=10$. For SAD-KL learning, we use the same $\beta=2500$ for all datasets and train the deep models for 300 epochs.

\begin{table}[b]
\vspace*{-5mm}
\caption{Average AUCs in $\%$ and training time for one vs. rest evaluation on MNIST dataset.}
\centering
\begin{tabular}{cccccccc}
\hline
\multirow{2}{*}{Normal class} & \multicolumn{2}{c}{Unsupervised learning} & \multicolumn{5}{c}{Semi-supervised learning}                                        \\
                              & SVDD                & Deep SVDD           & Deep SAD       & HSC            & SAD-PL (Hard)  & SAD-PL (Soft)   & SAD-KL         \\ \hline
0                             & 89.91               & 90.38               & 91.84          & 93.59          & 95.22          & 97.29           & 97.94          \\
1                             & 95.39               & 98.30               & 99.73          & 99.57          & 98.58          & 99.70           & 99.42          \\
2                             & 77.08               & 79.95               & 80.51          & 83.06          & 86.72          & 88.77           & 88.82          \\
3                             & 81.64               & 83.10               & 84.34          & 84.49          & 89.64          & 91.72           & 91.40          \\
4                             & 89.09               & 91.49               & 92.26          & 91.73          & 94.34          & 96.43           & 96.82          \\
5                             & 80.96               & 81.38               & 82.31          & 86.72          & 91.09          & 93.18           & 92.55          \\
6                             & 91.50               & 92.77               & 93.82          & 96.07          & 96.48          & 98.53           & 96.75          \\
7                             & 88.36               & 91.10               & 92.55          & 97.28          & 96.61          & 98.62           & 97.36          \\
8                             & 81.74               & 84.12               & 85.04          & 77.23          & 92.45          & 94.50           & 94.68          \\
9                             & 87.36               & 90.24               & 91.73          & 88.59          & 95.32          & 97.42           & 95.45          \\ \hline
Average                       & \textbf{86.30}      & \textbf{88.28}      & \textbf{89.41} & \textbf{89.83} & \textbf{93.65} & \textbf{95.62}  & \textbf{95.12} \\
Training time (s)             & \textbf{12.62}      & \textbf{36.45}      & \textbf{37.52} & \textbf{35.48} & \textbf{32.62} & \textbf{776.45} & \textbf{25.04} \\ \hline
\end{tabular}
\label{tab:table}
\vspace*{-5mm}
\end{table}

We use a typical one vs. rest evaluation method on the MNIST and CIFAR10 datasets [23]. On MNIST and CIFAR 10, we set the ten classes to be normal classes and let the remaining nine classes represent anomalies. We use the original training and test data. In the training data, we constitute most of the data as the normal class and replace a small amount with the data from the abnormal class according to the experimental scenario. We also divide the training data into labeled and unlabeled data, while organizing the abnormal samples in labeled data from a single anomalous class. However, the abnormal samples in unlabeled data equally contain samples of all anomalous classes. The class for abnormal samples in labeled data is randomly determined in each experiment. This gives training set sizes of approximately 6000 for MNIST and 5000 for CIFAR10. Both test sets have 10000 samples, including samples from the nine anomalous classes for each setup. On MNIST-C, we set original images to be normal and corrupted images to be abnormal. We use pre-configured training and test data. We also make the training data into most normal samples and a few abnormal samples according to the scenario. The training data are divided into 
labeled and unlabeled data. We organize the abnormal samples in labeled data using corrupted images of the same type. The abnormal samples in unlabeled data include all kinds of corrupted images equally. The type of corrupted images for abnormal samples in labeled data is randomly determined in each experiment. This configuration gives a training set size of approximately 60000 and a test set size of 160000 for MNIST-C. We use 5$\%$ as labeled data and 95$\%$ as unlabeled data in the training data as Ruff et al. [9] do for evaluation of the Deep SAD. We constituted 98$\%$ normal samples and 2$\%$ abnormalsamples in the labeled data, along with 99$\%$ normal samples and 1$\%$ abnormal samples in the unlabeled data. We present the evaluation results for the models with an average AUC of 30 times.

\begin{table}[]
\caption{Average AUCs in $\%$ and training time for one vs. rest evaluation on CIFAR10 dataset.}
\centering
\begin{tabular}{lccccccc}
\hline
\multicolumn{1}{c}{\multirow{2}{*}{Normal class}} & \multicolumn{2}{c}{Unsupervised learning} & \multicolumn{5}{c}{Semi-supervised learning}                                             \\
\multicolumn{1}{c}{}                              & SVDD                & Deep SVDD           & Deep SAD        & HSC             & SAD-PL (Hard)   & SAD-PL (Soft)    & SAD-KL          \\ \hline
Airplane                                          & 59.82               & 62.56               & 63.64           & 67.98           & 73.70           & 78.92            & 79.21           \\
Automobile                                        & 48.48               & 50.61               & 52.09           & 42.81           & 52.21           & 57.29            & 64.77           \\
Bird                                              & 61.28               & 62.63               & 64.41           & 64.39           & 58.41           & 63.82            & 64.62           \\
Cat                                               & 50.39               & 52.76               & 54.31           & 59.48           & 59.31           & 64.52            & 66.39           \\
Deer                                              & 64.73               & 65.96               & 68.37           & 72.89           & 77.35           & 82.64            & 77.83           \\
Dog                                               & 58.57               & 60.59               & 61.67           & 58.26           & 58.29           & 63.47            & 61.79           \\
Frog                                              & 67.04               & 69.20               & 70.32           & 68.31           & 71.68           & 76.89            & 73.38           \\
Horse                                             & 54.63               & 56.72               & 58.18           & 55.48           & 60.72           & 65.98            & 63.85           \\
Ship                                              & 64.33               & 66.89               & 68.29           & 76.78           & 69.35           & 74.52            & 73.87           \\
Truck                                             & 53.79               & 56.28               & 57.76           & 60.40           & 62.69           & 67.87            & 73.47           \\ \hline
\multicolumn{1}{c}{Average}                       & \textbf{58.30}      & \textbf{60.42}      & \textbf{61.90}  & \textbf{62.68}  & \textbf{64.37}  & \textbf{69.59}   & \textbf{69.91}  \\
\multicolumn{1}{c}{Training time (s)}             & \textbf{70.27}      & \textbf{186.45}     & \textbf{188.52} & \textbf{184.48} & \textbf{179.67} & \textbf{1436.67} & \textbf{167.97} \\ \hline
\end{tabular}
\label{tab:table}
\end{table}

\begin{table}
\begin{minipage}[b]{0.45\linewidth}
\centering
\caption{Average AUCs in $\%$ and training time for evaluation on MNIST-C dataset.}
\begin{tabular}{llcc}
\hline
\multicolumn{2}{l}{Model}                                                                              & AUC   & \begin{tabular}[c]{@{}c@{}}Training\\ time (s)\end{tabular} \\ \hline
\multirow{2}{*}{\begin{tabular}[c]{@{}l@{}}Unsupervised\\ learning\end{tabular}}       & SVDD          & 67.59 & 2970                                                        \\
                                                                                       & Deep SVDD     & 82.79 & 7927                                                        \\
\multirow{5}{*}{\begin{tabular}[c]{@{}l@{}}Semi-\\ supervised\\ learning\end{tabular}} & Deep SAD      & 89.89 & 7945                                                        \\
                                                                                       & HSC           & 85.10 & 7894                                                        \\
                                                                                       & SAD-PL (Hard) & 90.05 & 7562                                                        \\
                                                                                       & SAD-PL (Soft) & 92.43 & 83999                                                       \\
                                                                                       & SAD-KL        & 95.26 & 6192                                                        \\ \hline
\end{tabular}
\label{tab:table}
\vspace*{+10mm}
\end{minipage}\hfill
\begin{minipage}[b]{0.45\linewidth}
\centering
\includegraphics[width=75mm]{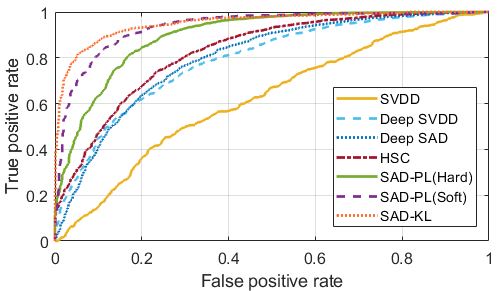}
\captionof{figure}{The ROC curve obtained from MNIST-C dataset. }
\label{fig:image}
\end{minipage}
\vspace*{-7.5mm}
\end{table}

Table 1 shows the evaluation results using MNIST dataset. The SAD-KL achieves a better performance than the unsupervised and semi-supervised anomaly detection models except for SAD-PL(Soft) with an average AUC of 95.12$\%$. However, note that SAD-KL has 31 times faster training with little performance degradation of 0.50$\%$ and is at least 7.25sec faster than other deep anomaly detection models. Tables 2 shows the evaluation results using CIFAR10 dataset. Similarly, the SAD-KL shows better performance than existing anomaly detection models with an average AUC of 69.91$\%$ and fastest training time. Tables 3 shows anomaly detection evaluation results using MNIST-C dataset. The SAD-KL achieves the best performance over all unsupervised and semi-supervised models with an average AUC of 95.26$\%$ and faster training time than the semi-supervised models.

Figure 6 shows the receiver operation characteristic (ROC) curve obtained from MNIST-C dataset which contains anomalous images of multimodal normality. Note that the SAD-KL shows the best ROC performance even the unlabeled data has diverse statistics. We present anomaly scores of normal and abnormal images obtained from 4 anomaly detection models in Figure 7. We can observe that anomaly scores of the SAD-KL are the biggest with abnormal image and compatible to the SAD-PL(Soft) with normal image.

Next, we investigate the effect of including labeled data during training on the MNIST-C datasets by increasing the ratio of labeled training data from 0$\%$ to 10$\%$ and presenting the averaged AUCs of 30 times. The ratio of abnormal samples in labeled and unlabeled data is maintained at 5$\%$ and 1$\%$, as in the previous experiments. Figure 8 (a) shows variations of performance in the average AUCs for the semi-supervised models according to the ratio of labeled data during training. Note that the SAD-KL achieve the best AUC performance even with a high proportion of labeled data during training.
Similarly, we investigate the effect of including abnormal samples in the unlabeled data during the training with the MNIST-C datasets. To do this, we increase the anomaly ratio, which is the proportion of abnormal samples in unlabeled training data, from 0$\%$ to 10$\%$ and represent the average AUCs of 30 times. For the experiment, we keep the labeled data at 95$\%$ of the training data and include 98$\%$ and 2$\%$ of normal and abnormal samples, respectively. Figure 8 (b) shows the variations of performance for the average AUCs according to the anomaly ratio in unlabeled data. Note than existing unsupervised and semi-supervised algorithms show performance degradation in the average AUCs as the anomaly ratio in unlabeled data increases. However, the SAD-KL shows improved AUC performance as the anomaly ratio in unlabeled data increases, which implies that the proposed KL divergence-based detection algorithm is robust to ratio of normal data to abnormal data.

\begin{figure}
  \centering
  \includegraphics[width=\textwidth]{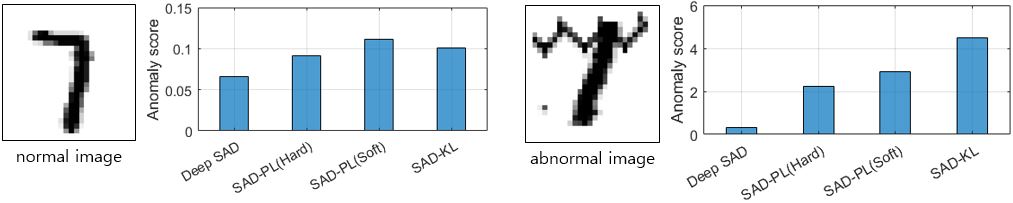}
  \vspace*{-5mm}
  \caption{Normal and abnormal images along with their anomaly scores.}
  \label{fig:fig7}
\end{figure}

\begin{figure}
  \centering
  \includegraphics[width=\textwidth]{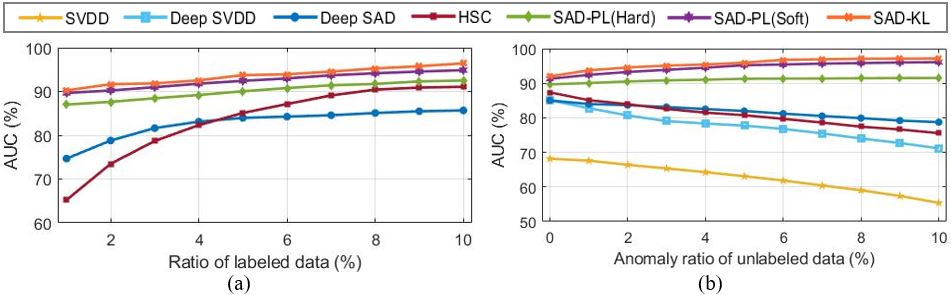}
  \vspace*{-5mm}
  \caption{Average AUCs according to (a) ratio of labeled data and anomaly (b) ratio of unlabeled data for evaluation using MNIST-C dataset.}
  \label{fig:fig8}
  \vspace*{-5mm}
\end{figure}

\section{Conclusion}
To detect normal data in unlabeled data, we propose KL divergence-based algorithm. The proposed SAD-KL uses KL divergence of the PDFs of LOF scores obtained from both labeled and unlabeled data and then estimates the threshold for normal data detection. The SAD-KL takes less learning time than existing semi-supervised anomaly detection algorithms since it estimates PDFs of the data and then evaluate the detection threshold from the computed CDF. Through experiments, we show that PDFs of LOFs follow Burr distribution and that the SAD-KL presents a higher performance in the average AUCs, displays tighter decision boundaries and achieves higher learning efficiency than the existing algorithms. Therefore, the proposed SAD-KL can be a strong candidate for efficient semi-supervised learning algorithm requiring stable detection performance regardless of distribution variation between the labeled and unlabeled data. The inclusion of labeled abnormal data in detection probability estimation and the extension of the SAD-KL to explainable anomaly detection become interesting topics for future research.

\newpage

\section{References}

[1] L. Ruff, J. R. Kauffmann, R. A. Vandermeulen, G. Monatavon, W. Samek, M. Kloft, T. G. Dietterich, and K. R. Müller. A unifying review of deep and shallow anomaly detection. Proceeding of the IEEE, 109(5):756-795, 2021. [2] B. Schölkopf, R. Willianson, A. Smola, J. S. Taylor, and J. 850 Platt. Support vector method for novelty detection. In NIPS, 851 1999.

[3] D. M. Tax, and R. P. Duin. Support vector data description. Machine learning, 59(1):45-66, 2004 .

[4] L. Ruff, R. A. Vandermeulen, N. Görnits, L. Deecke, S. A. Siddiqui, A. Binder, E. Müller, and M. Kloft. Deep one-class classification. In ICML, 2018.

[5] Z. Ghafoori, and C. Leckie. Deep multi-sphere support vector data description. In SIAM, $2020 .$

[6] S. Goyal, A. Raghunathan, M. Jain, H. Simhadri, and P. Jain. DROCC: Deep robust one-class classification. In ICML, 2020.

[7] H. Song, Z. Jiang, A. Men, and B. Yang. A hybrid semisupervised anomaly detection model for high-dimensional data. Computational Intelligence and Neuroscience, vol. 2017.

[8] S. Akvay, A. A. Abarghouel, and T. P. Breckon. GANomaly: semi-supervised anomaly detection via adversarial training. In ACCV, 2018.

[9] L. Ruff, R. A. Vandermeulen, N. Görnitz, A. Binder, E. Müller, K. R. Müller, and M. Kloft. Deep semi-supervised anomaly detection. In ICLR, 2020.

[10] D. Hendrycks, M. Mazeika, and T. Dietterich. Deep anomaly detection with outlier exposure. In ICLR, 2019.

[11] L. Ruff, R. A. Vandermeulen, B. J. Franks, K. R. Müller, and M. Kloft. Rethinking assumptions in deep anomaly detection. In ICML, 2021.

[12] K. Lee, C. H. Lee, and J. Lee. Semi-supervised anomaly detection algorithm using probabilistic labeling (SAD-PL). IEEE Access, 9:142972-142981, 2021.

[13] H. Ju, D. Lee, J. Hwang, J. Namkung, and H. Yu. PUMAD: PU metric learning for anomaly detection. Information Sciences, 523:167-183, 2020.

[14] L. Yang, J. Chen, Z. Wang, W. Wang, J. Jiang, X. Dong, and W. Zhang. PLELog: Semi-supervised log-based anomaly detection via probabilistic label estimation. In ICSE, 2021.

[15] L. McInnes, J. Healy, and S. Astels. hdbscan: Hierarchical density based clustering. J. Open Source Softw., 2(11):205, 2017.

[16] M. M. Breuning, H. P. Kriegl, R. T. Ng, and J. Sander. LOF: Identifying density-based local outliers. In ACM SIGMOD, 2000.

[17] J. Neyman, and E. S. Pearson. On the problem of the most efficient tests of statistical hypotheses. Philosophical Transactions of the Royal Society of London. Series A, 231:289-337, 1933.

[18] I. W. Burr. Cumulative frequency functions. Annals of Mathematical Statistics. 13(2):215-232, 1942.

[19] A. R. Hakim, I Fithriani, and M. Novita. Properties of Burr distribution and its application to heavy-tailed survival time data. J. Phys.: Conf. Ser., 2021.

[20] Y. Lecun, C. Cortes, and C. Burges. MNIST handwritten digit database. AT\&T Labs, 2010.

[21] A. Krizhevsky. Learning multiple layers of features from tiny images. 2009.

[22] N. Mu, and J. Gilmer. MNIST-C: robustness benchmark for computer vision. In ICML, 2019.

[23] A. F. Emmott, S. Das, T. Dietterich, A. Fern, and W. K. Wong. Systematic construction of anomaly detection bechmarks from real data. In ACM SIGKDD, 2013.

\newpage

\section*{A. Validation for Burr distribution assumption}
We plot histograms, the estimated PDFs and the CDFs of $s_{u}$ of the MNIST, CIFAR10, and MNIST-C datasets in Figure S1, which validates the Burr distribution assumption.

\begin{figure}[h]
  \centering
  \captionlabelfalse
  \includegraphics[width=\textwidth]{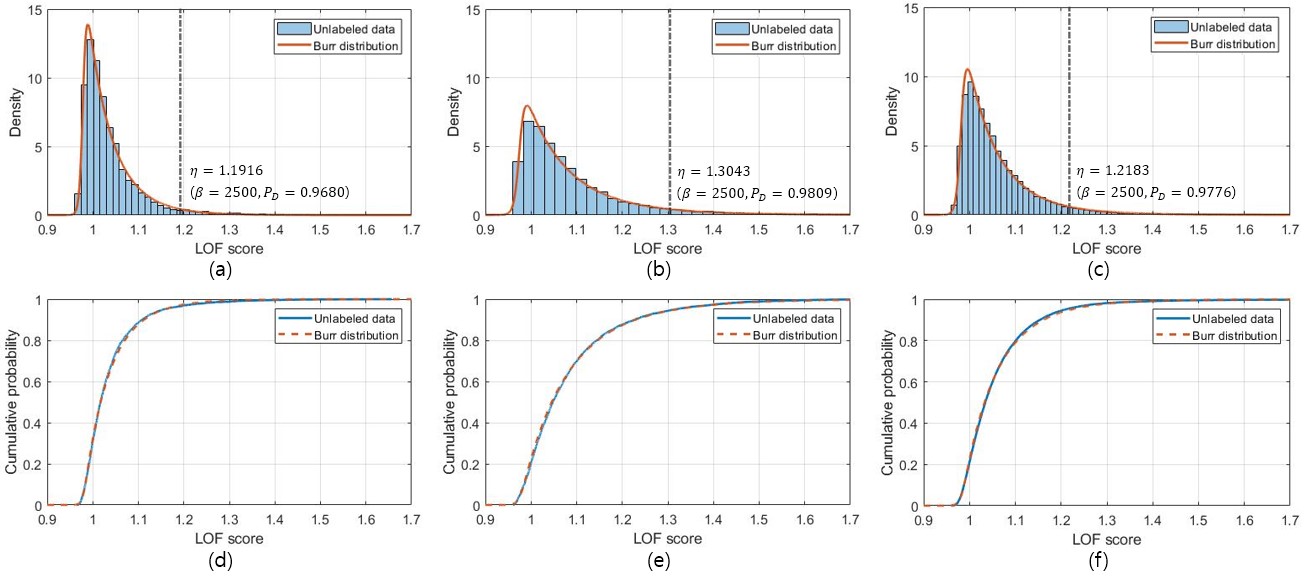}
  \vspace*{-5mm}
  \caption{Figure S1: Histogram and the Burr PDF obtained from (a) MNIST (b) CIFAR10 (c) MNIST-C and the estimated CDF and the Burr CDF from (d) MNIST (e) CIFAR10 (f) MNIST-C dataset.}
  \addtocounter{figure}{-1} 
\end{figure}

We did Kolmogorov-Smirnov (KS) test and two histogram-based tests on MNIST-C and summarize results in Table S1, which validates our assumption

\begin{table}[h]
\captionlabelfalse
\caption{Table S1: Validation tests on MNIST-C dataset.}
\centering
\begin{tabular}{ccc}
\hline
KS-Test ($\alpha=0.05$) & Histogram Intersection & Histogram Correlation \\ \hline
p-value$=0.9855>\alpha$  & 0.9682                 & 0.9981                \\ \hline
\end{tabular}
\addtocounter{table}{-1}
\end{table}

\section*{B. 	Estimation of parameter $\mathbf{\beta}$}
We estimated KL divergences of four datasets and summarize them in Table S2.

\begin{table}[h]
\captionlabelfalse
\caption{Table S2: KL divergences of four datasets.}
\centering
\begin{tabular}{cccl}
\hline
MNIST & CIFAR10 & MNIST-C & Two moons \\ \hline
81.30 & 48.21   & 60.73   & 20.44     \\ \hline
\end{tabular}
\addtocounter{table}{-1}
\end{table}

From $P_{D}=\exp (-K L(P \| Q) / \beta)$, we can obtain equation for $\beta$ as follows:

\begin{equation} \tag{1}
\beta=-\mathrm{KL}(\mathrm{P} \| \mathrm{Q}) / \log \left(P_{D}\right)
\end{equation}

The $\beta$ can be determined once range of $P_{D}$ is chosen a priori as shown in Figure S2. Note that higher $P_{D}$ does not give higher AUC since the $P_{D}$ is for LOF detection not for anomaly detection. Suppose the $P_{D}$ ranges from 0.95 to 0.98, then $\beta$ ranges from 1600 to 3000 for MNIST. In supplementary material we showed $\beta \approx 2500$ seems to be appropriate for three datasets. When other dataset is considered, range of $P_{D}$ and subsequent choice of $\beta$ still can be estimated by equation (1) using new KL divergence.

We present the detection performance of the SAD-KL with respect to the $\beta$ by changing it from 2000 to 3000 with 50 increasement. The results shown in Figure S2 prove that the SAD-KL’s AUC performance is fairly stable in wide range of the $\beta$ chosen for three datasets.

\section*{C.	Detection performance according to training data ratio }

We investigate the AUCs obtained by changing percentage of the labeled data and percentage of the abnormal data in the MNIST and the CIFAR10 datasets. The results are presented in Figure S3, which shows the performance of the proposed SAD-KL is compatible to the SAD-PL requiring heavy computation.

\begin{table}[h]
\begin{minipage}[b]{0.48\linewidth}
\centering
\captionlabelfalse
\includegraphics[width=80mm]{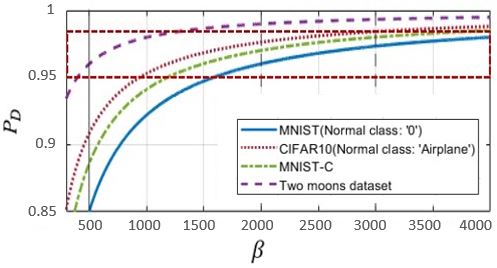}
\vspace*{-2mm}
\captionof{figure}{Figure S2: $P_{D}$ according to $\beta$.}
\addtocounter{figure}{-1}
\end{minipage}\hfill
\begin{minipage}[b]{0.48\linewidth}
\centering
\captionlabelfalse
\includegraphics[width=78mm]{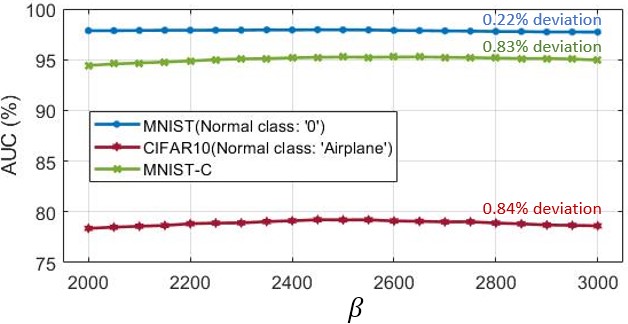}.
\vspace*{-2mm}
\captionof{figure}{Figure S3: The AUCs according to $\beta$ on the MNIST, CIFAR10, and MNIST-C datasets.}
\addtocounter{figure}{-1}
\end{minipage}
\vspace*{-3mm}
\end{table}

\section*{D. Training time of the SAD-KL}
While the SAD-PL adopts ensemble networks to choose the best network, the SAD-KL uses single network with fast threshold computation, which yields fast training. Even the SAD-KL takes little extra computation of LOF every epoch, due to KL divergence-based labeling it shows faster convergence than the Deep SAD and HSC. The superior convergence performance is shown in Figure S4.

\begin{figure}[h]
  \centering
  \captionlabelfalse
  \includegraphics[width=0.5\textwidth]{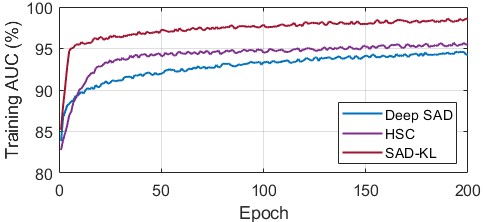}
  \vspace*{-1mm}
  \caption{Figure S4: AUC of MNIST-C according to epoch.}
  \addtocounter{figure}{-1} 
  \vspace*{-3mm}
\end{figure}

\section*{E. Comparisons with AnoGAN and GANomaly}
We performed experiments by using 1 class being “abnormal” and the others being “normal” as the AnoGAN and GANomaly do. The AUC results shown in Figure S5 shows superior AUC performance of the SAD-KL.

\begin{figure}[h]
  \centering
  \captionlabelfalse
  \includegraphics[width=0.75\textwidth]{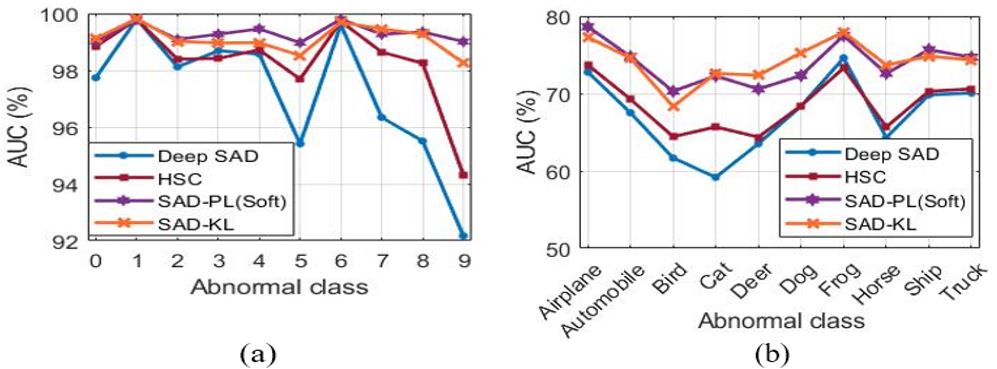}
  \vspace*{-2mm}
  \caption{Figure S5: AUC results on (a)MNIST and (b)CIFAR10 s.}
  \addtocounter{figure}{-1} 
\end{figure}

\section*{F.	Anomaly score comparisons with normal and abnormal images}
We present normal and abnormal test images along with the corresponding anomaly scores of the SAD-KL and four models in Figure S6 and S7. The anomaly scores are computed as $\Vert\phi(x)-\mathbf{c}\Vert^{2}$ for Deep SAD, SAD-PL and SAD-KL and as $\Vert\phi(x)\Vert^{2}$ for HSC. Note the SAD-KL’s scores are bigger in most MNIST images and is the biggest in MNIST-C images.

\begin{figure}[h]
  \centering
  \captionlabelfalse
  \includegraphics[width=0.9\textwidth]{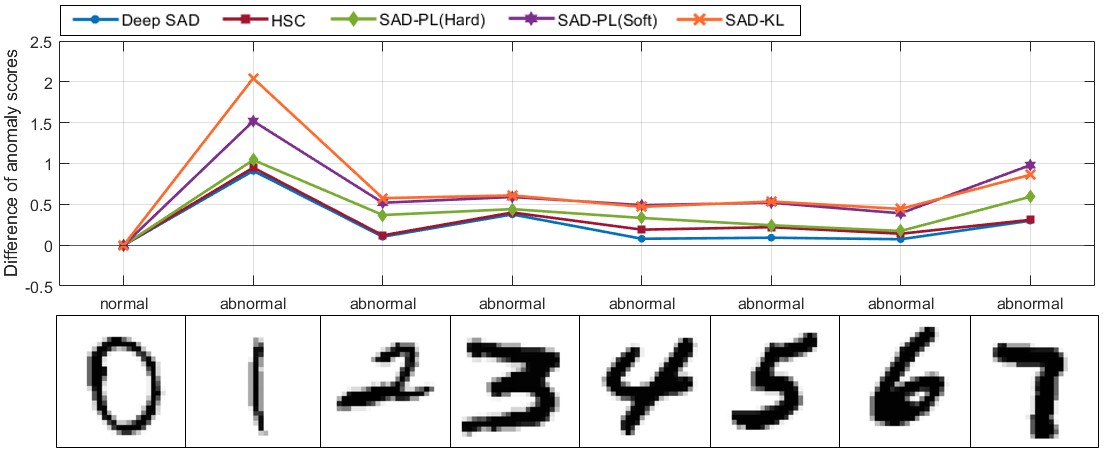}
  \vspace*{-2mm}
  \caption{Figure S6: Anomaly scores of test images in MNIST dataset and the corresponding images.}
  \addtocounter{figure}{-1}
\end{figure}

\begin{figure}[h]
  \centering
  \captionlabelfalse
  \includegraphics[width=0.9\textwidth]{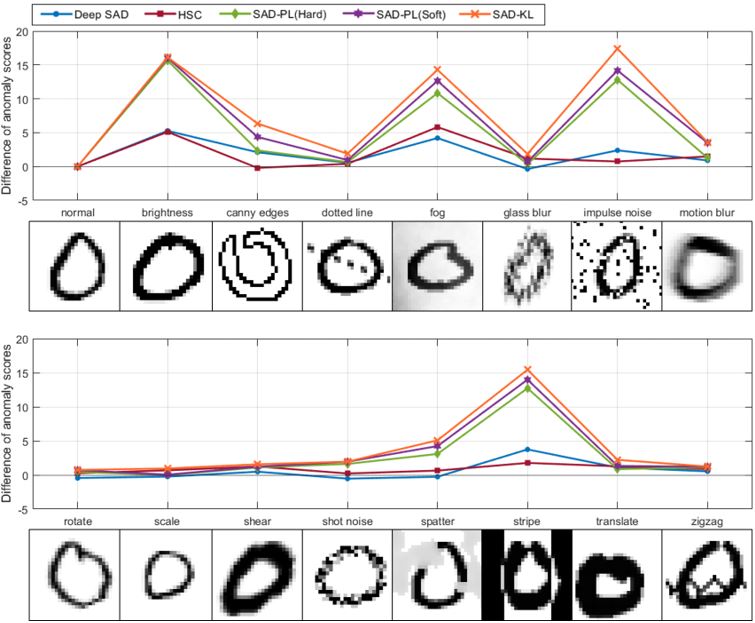}
  \vspace*{-2mm}
  \caption{Figure S7: Anomaly scores of test images in MNIST-C dataset and the corresponding images.}
  \addtocounter{figure}{-1} 
  \vspace*{-10mm}
\end{figure}

\end{document}